\documentclass{iise}
\usepackage{tcolorbox}
\usepackage{multirow}
\usepackage{booktabs}
\usepackage{graphicx}
\usepackage{subfig}

\conference{Proceedings of the IISE Annual Conference \& Expo 2025 \\
E. Gentry, F. Ju, X. Liu, eds.}

\title{\titlesize Leveraging Large Language Models for Risk Assessment in Hyperconnected Logistic Hub Network Deployment}

\author{
Yinzhu Quan\textsuperscript{1}, Yujia Xu\textsuperscript{1}, Guanlin Chen\textsuperscript{1}, Frederick Benaben\textsuperscript{1,2}, and Benoit Montreuil\textsuperscript{1}\\
\textsuperscript{1}Physical Internet Center, H. Milton Stewart School of Industrial and Systems Engineering, Georgia Institute of Technology, Atlanta, GA, USA \\
\textsuperscript{2}Industrial Engineering Centre, IMT Mines Albi, Albi, France
}

\authorlist{Quan, Xu, Chen, Benaben, and Montreuil}
\abstractID{6530}

\begin{document}
\maketitle

\begin{abstract}
{\small The growing emphasis on energy efficiency and environmental sustainability in global supply chains introduces new challenges in the deployment of hyperconnected logistic hub networks. In current volatile, uncertain, complex, and ambiguous (VUCA) environments, dynamic risk assessment becomes essential to ensure successful hub deployment. However, traditional methods often struggle to effectively capture and analyze unstructured information. In this paper, we design an Large Language Model (LLM)-driven risk assessment pipeline integrated with multiple analytical tools to evaluate logistic hub deployment. This framework enables LLMs to systematically identify potential risks by analyzing unstructured data, such as geopolitical instability, financial trends, historical storm events, traffic conditions, and emerging risks from news sources. These data are processed through a suite of analytical tools, which are automatically called by LLMs to support a structured and data-driven decision-making process for logistic hub selection. In addition, we design prompts that instruct LLMs to leverage these tools for assessing the feasibility of hub selection by evaluating various risk types and levels. Through risk-based similarity analysis, LLMs cluster logistic hubs with comparable risk profiles, enabling a structured approach to risk assessment. In conclusion, the framework incorporates scalability with long-term memory and enhances decision-making through explanation and interpretation, enabling comprehensive risk assessments for logistic hub deployment in hyperconnected supply chain networks.}
\end{abstract}

\section*{Keywords}
Risk Assessment, Large Language Models, Supply Chain Resilience, LLM-Driven Pipeline, Risk-Based Clustering, Logistic Hub Networks, Physical Internet

\section{Introduction}

Hyperconnected logistics hub networks serve as critical enablers of open freight flow consolidation and seamless asset and resource sharing across different parties and transportation modes. These hubs play a pivotal role in aggregating smaller shipments into larger, more energy-efficient loads, thereby reducing costs and improving operational efficiency. Furthermore, hubs act as strategic nodes that facilitate collaboration among stakeholders, ensuring smooth coordination and adaptability. Figure~\ref{fig:network} illustrates the complexity and connectivity of such networks, highlighting their role in streamlining logistics operations and enhancing resilience. By integrating advanced analytics with these networks, organizations can optimize hub placement and operations, ensuring they are both sustainable and resilient to disruptions.

Risk assessment is a cornerstone of resilient supply chain and logistics engineering and management, especially in the context of hyperconnected logistics networks. To enhance energy efficiency and sustainability in global supply chains, the design and deployment of logistic hub networks \cite{montreuil2018urban} must address various types of risks, such as environmental, geopolitical, and operational risks \cite{arowosegbe2024sustainability}. These risks are further magnified in volatile, uncertain, complex, and ambiguous (VUCA) environments, where failures in hub location deployment can lead to significant financial and operational setbacks. Effective risk assessment ensures not only the reliability of logistics operations but also the resilience of the entire supply chain networks. Integrating advanced analytics into risk assessment is essential as traditional methods fail to address the complexity of modern logistics \cite{wang2016big}. 

Traditional risk assessment methods, while systematic, often lack the capacity to process unstructured data such as historical weather events or real-time geopolitical updates, limiting their applicability in dynamic scenarios. Machine learning (ML) approaches have addressed some of these limitations by enabling predictive analyses and decision-making based on historical trends and large datasets \cite{boppiniti2019machine}. However, these methods frequently require substantial data preprocessing and are often constrained by their reliance on structured data formats. Recently, large language models (LLMs) have been increasingly applied in the supply chain domain \cite{quan2024econlogicqa, quan2024invagent}, offering unparalleled capabilities for analyzing unstructured data and generating insights across diverse applications. By leveraging LLMs, our approach integrates multiple tools to automate risk analysis, providing a scalable, adaptive, and context-aware framework for hub placement in logistic networks. This integration facilitates the efficient evaluation of logistic hubs across combinations of regions and time periods while enhancing interpretation, effectively addressing the complexities of hyperconnected logistics networks.

The main contributions of this paper are as follows:
\begin{enumerate}
    \item \textbf{Integration of Multiple Tools:} We propose an LLM-driven risk assessment pipeline that integrates multiple analytical tools to enhance logistic hub deployment decisions. This pipeline enables LLMs to systematically process both structured and unstructured data by leveraging hub databases, Wikipedia-based contextual insights, financial analysis, storm event records, news aggregation, and traffic monitoring.
    
    \item \textbf{Scalability and Interpretability:} Our framework leverages LLMs' long-term memory to efficiently process extensive datasets across multiple regions and time periods, ensuring scalability. Additionally, it enhances interpretability by providing explanations of risk types, severity levels, and their impact on hub selection, fostering transparency and insight-driven risk management.
    
    \item \textbf{Risk-Based Clustering:} By employing risk-based similarity analysis, our approach clusters logistic hubs with comparable risk profiles. This clustering mechanism provides insights for risk management, resource allocation efficiency, and proactive mitigation strategies.
\end{enumerate}

\section{Related Work}

Risk management in supply chain logistics is critical for ensuring resilience in hyperconnected networks. Traditional frameworks emphasize systematic risk identification and mitigation \cite{tummala2011assessing}, while recent advancements highlight the role of technology in addressing interconnected risks \cite{choudhary2023risk}. Emerging research underscores the integration of artificial intelligence (AI) and decision-making models for adaptive risk assessment \cite{emrouznejad2023supply} and stresses the importance of proactive strategies to manage disruptions like natural disasters and pandemics \cite{pournader2020review}. These insights form the foundation for leveraging large language models (LLMs) to analyze risks in dynamic and data-intensive logistics environments.

Building on this foundation, machine learning (ML) and AI have further enhanced supply chain risk management (SCRM) by offering predictive capabilities and improving resilience. \citet{aljohani2023predictive} highlights ML's role in proactive risk assessment through real-time monitoring, while \citet{yang2023supply}  emphasize the use of advanced algorithms like Random Forest to predict and mitigate disruptions. \citet{jahin2023ai} underscore ML's adaptability in enhancing resilience and managing cascading risks, and \citet{nezianya2024critical} review their potential for improving overall risk identification and assessment strategies in supply chains. These advancements support leveraging LLMs for synthesizing unstructured data and enabling robust risk assessment in hyperconnected logistic hubs.

LLMs revolutionize SCRM by enabling efficient analysis of unstructured data. \citet{zhao2024optimizing} demonstrate automated risk identification and categorization to enhance resilience, while \citet{sun2024application} highlight LLMs’ superiority in early risk detection through news analysis. These advancements showcase LLMs’ potential for real-time, scalable risk assessments, supporting their application in hyperconnected logistic hub deployment.

\section{Problem Definition}

The deployment of logistic hub networks in global supply chains is filled with risks such as environmental hazards and operational disruptions. Traditional risk assessment methods are limited in their capacity to process unstructured data, which is vital for understanding factors like historical storm events and geopolitical changes. These limitations hinder the effective evaluation of hub feasibility and resilience. In this paper, we conduct risk assessment for logistics
hubs proposed for a Physical Internet  based hyperconnected logistic network \cite{montreuil2011toward, kulkarni2022resilient, li2024stochastic, muthukrishnanresilient, xu5144739dynamic} designed based on data extracted from Freight Analysis Framework (FAF) database produced by the Bureau of Transportation Statistics \footnote{\url{https://www.bts.gov/faf} (Last accessed on February 8, 2025)} (BTS). Each logistic hub is associated with a set of risks, and conversely, each risk impacts a set of hubs. The consideration of each logistic hub is based solely on its geographical state, latitude, and longitude. We use Georgia as the testbed in this study. In summary, assuming all hubs are deployed, this research aims to leverage large language models (LLMs) and external tools to analyze both structured and unstructured data, identify risk types, conduct daily risk assessments and aggregate them into yearly summaries, and cluster logistic hubs based on their risk profiles to uncover regional risk patterns and support strategic decision-making in logistic hub deployment.

\section{Methodology}

\begin{figure}[h!]
    \centering
    \begin{minipage}[t]{0.3\linewidth}
        \centering
        \includegraphics[height=0.25\textheight]{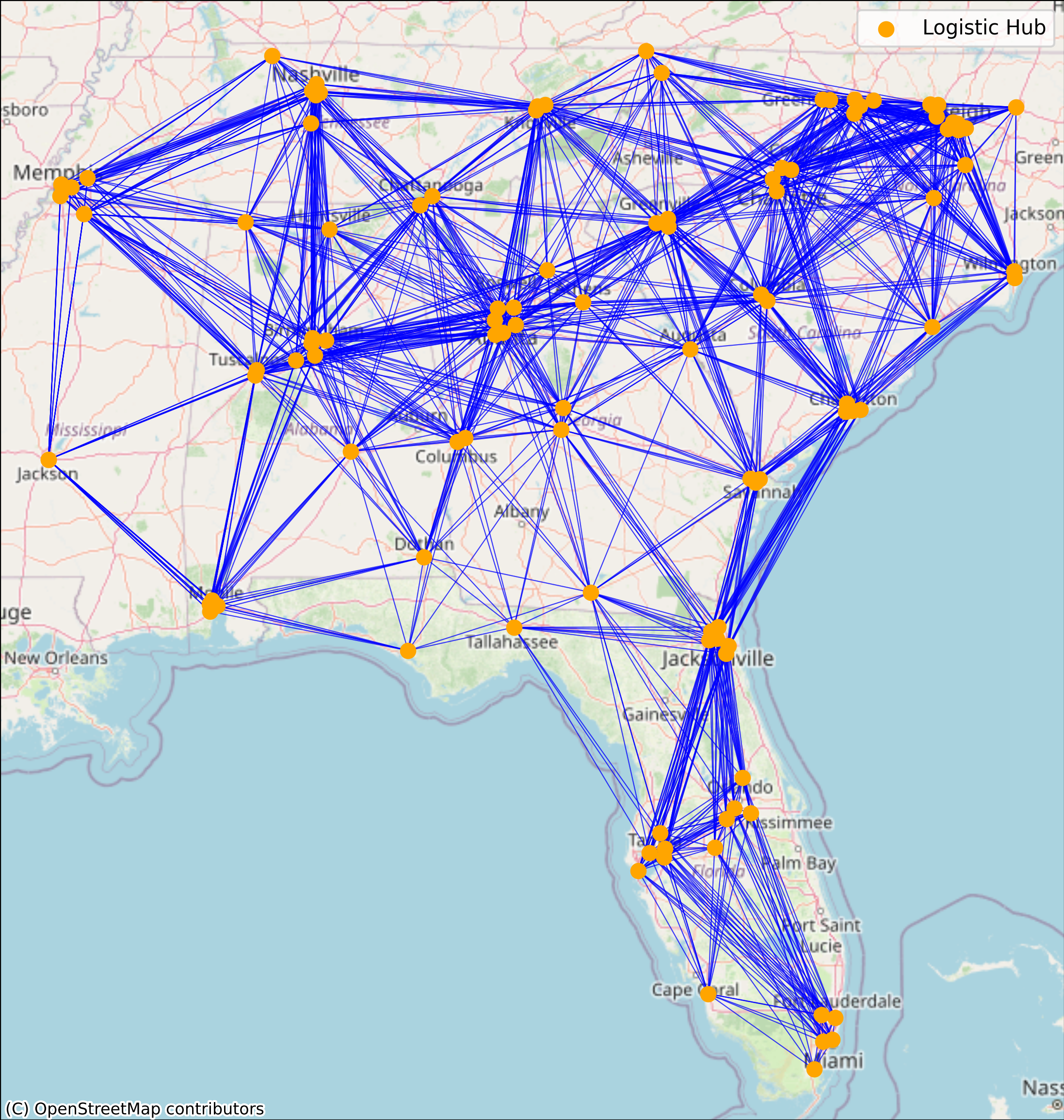}
        \caption{Hyperconnected logistic hub networks.}
        \label{fig:network}
    \end{minipage}
    \hfill
    \begin{minipage}[t]{0.65\linewidth}
        \centering    \includegraphics[height=0.25\textheight]{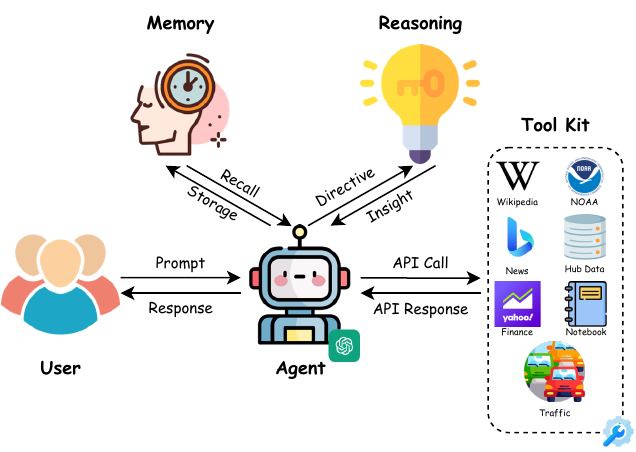}
        \caption{Workflow of the multi-tool integrated framework for risk assessment in logistic hub deployment.}
        \label{fig:flowchart}
    \end{minipage}
\end{figure}

We present a framework that integrates large language models (LLMs) with a suite of external tools to conduct comprehensive risk assessments for logistic hub deployment. The workflow of the multi-tool integrated framework for risk assessment in logistic hub deployment, as shown in Figure~\ref{fig:flowchart}, is designed to enable seamless and dynamic risk analysis. The process begins when the user defines tasks or directives, such as analyzing risks for logistic hubs or specifying time intervals. These prompts are then processed by the central agent, which orchestrates multiple tools, memory, and reasoning capabilities to generate insights. The toolkit comprises specialized tools, each designed to process different types of structured and unstructured data and contribute to comprehensive risk assessment. Memory retains context for scalable, iterative analysis, and reasoning synthesizes outputs for actionable insights.

To effectively implement this workflow, we integrate a suite of external tools that enhance LLMs’ ability to systematically evaluate risk factors at different stages of the assessment process. The Hub Information Extraction Tool processes structured logistic hub data, extracting key attributes to ensure smooth integration with the risk assessment framework. The Wikipedia Summary and Source Tool provides contextual insights on geopolitical, cultural, and environmental factors, supporting a comprehensive regional risk analysis. The Financial Data Analysis Tool evaluates economic trends and stability to assess the financial viability of logistic hubs. The NOAA Storm Data Tool retrieves historical storm data, including event type, property damage, injuries, and fatalities, to evaluate natural disaster risks in hub locations. The News Aggregation Tool monitors and summarizes emerging risks, such as political instability or regulatory changes, using real-time news updates. The Traffic Status Tool provides real-time congestion data, enabling the assessment of traffic conditions and operational risks around logistic hubs. Finally, to optimize processing efficiency, the Notebook Tool manages LLM token limits, processes data in structured batches, and systematically organizes results into a final summary.

The agent dynamically invokes these tools based on user instructions and data requirements, while LLMs synthesize structured and unstructured data into actionable insights. The framework produces detailed risk assessments, feasibility classifications, and explanations for hub placement decisions. The memory component retains short-term context within multi-turn interactions, while the Notebook Tool serves as long-term storage, ensuring scalability and consistency. By leveraging long-term memory, the system can track evolving risk patterns and incorporate historical data into future assessments. This enhances the adaptability of the framework, enabling it to refine recommendations as new risks emerge and operational demands shift. Ultimately, this interconnected system provides a robust foundation for risk-aware logistic hub deployment in dynamic environments.

In our current implementation, we design and implement a \texttt{ReAct}-based \cite{yao2022react} framework tailored for dynamic decision-making in supply chain operations, leveraging the \texttt{LangGraph} \footnote{\url{https://langchain-ai.github.io/langgraph/} (Last accessed on February 8, 2025)} platform to integrate reasoning and action. Using \texttt{ChatOpenAI} \footnote{\url{https://python.langchain.com/docs/integrations/chat/openai/} (Last accessed on February 8, 2025)} with the \texttt{gpt-4o} model, we initialize a reasoning engine that interleaves logical inference with task-specific actions. This agent-driven system, built with \texttt{LangGraph}, dynamically connects the language model’s reasoning capabilities with external tools for efficient data retrieval and analysis. By facilitating iterative workflows, the framework processes diverse data sources, extracts actionable insights, and adapts decisions in real time. With its scalability, flexibility, and precision, the \texttt{ReAct} framework effectively addresses the complexities of modern supply chain environments.

\section{Case Study}

A structured two-step approach is implemented to assess risks associated with logistic hub deployment. The first step focuses on identifying risk types and selecting the most effective tools for assessment. The second step involves conducting daily risk assessments for each logistic hub, aggregating the results into yearly summaries, and analyzing the relationship between hubs and identified risks.

To evaluate the framework's ability to identify risk types (see Figure~\ref{fig:prompt_identify_risks} for an example response) associated with logistic hubs, we conduct experiments using multiple representative time intervals across different seasons. These periods are selected to balance computational efficiency and prompt limitations while ensuring a comprehensive evaluation of risk patterns throughout the year. In this step, the focus is on identifying the most effective tools for extracting relevant risk information rather than evaluating the severity levels of risks. The analysis highlights the most frequently occurring risk types, which include thunderstorm winds, tornadoes, flash floods, lightning, floods, hail, strong winds, and traffic jams. To further refine the analysis and strengthen the connection between hubs and risks, we select the NOAA Storm Data Tool, Wikipedia Summary and Source Tool, and Traffic Status Tool as the most effective tools for this assessment.

\begin{figure*}[h!]
  \centering
  \begin{tcolorbox}
    
    \textbf{Response:}

    {\small - Hub Location: Latitude 33.712066, Longitude -84.236439, State: Georgia\\
    - Risks:\\
    \hspace*{2mm} - Traffic Jam\\
    - Explanation: The traffic at the location (coordinates: 33.712066,-84.236439) is currently severe traffic jam (long delay). The current speed is 19 km/h, compared to a normal free-flow speed of 90 km/h. Vehicles are expected to take approximately 107 seconds to pass through this segment, which is 85 seconds longer than the typical free-flow travel time of 22 seconds. The data confidence level is 1.0 out of 1, ensuring the reliability of this information.}
    
  \end{tcolorbox}
  \caption{An example response to identify risk types for logistic hubs in Georgia using Traffic Status Tool.}
  \label{fig:prompt_identify_risks}
\end{figure*}

\begin{figure}[h!]
    \centering
    \subfloat[Risk-based bub similarity matrix.]{\includegraphics[height=0.29\textheight,keepaspectratio]{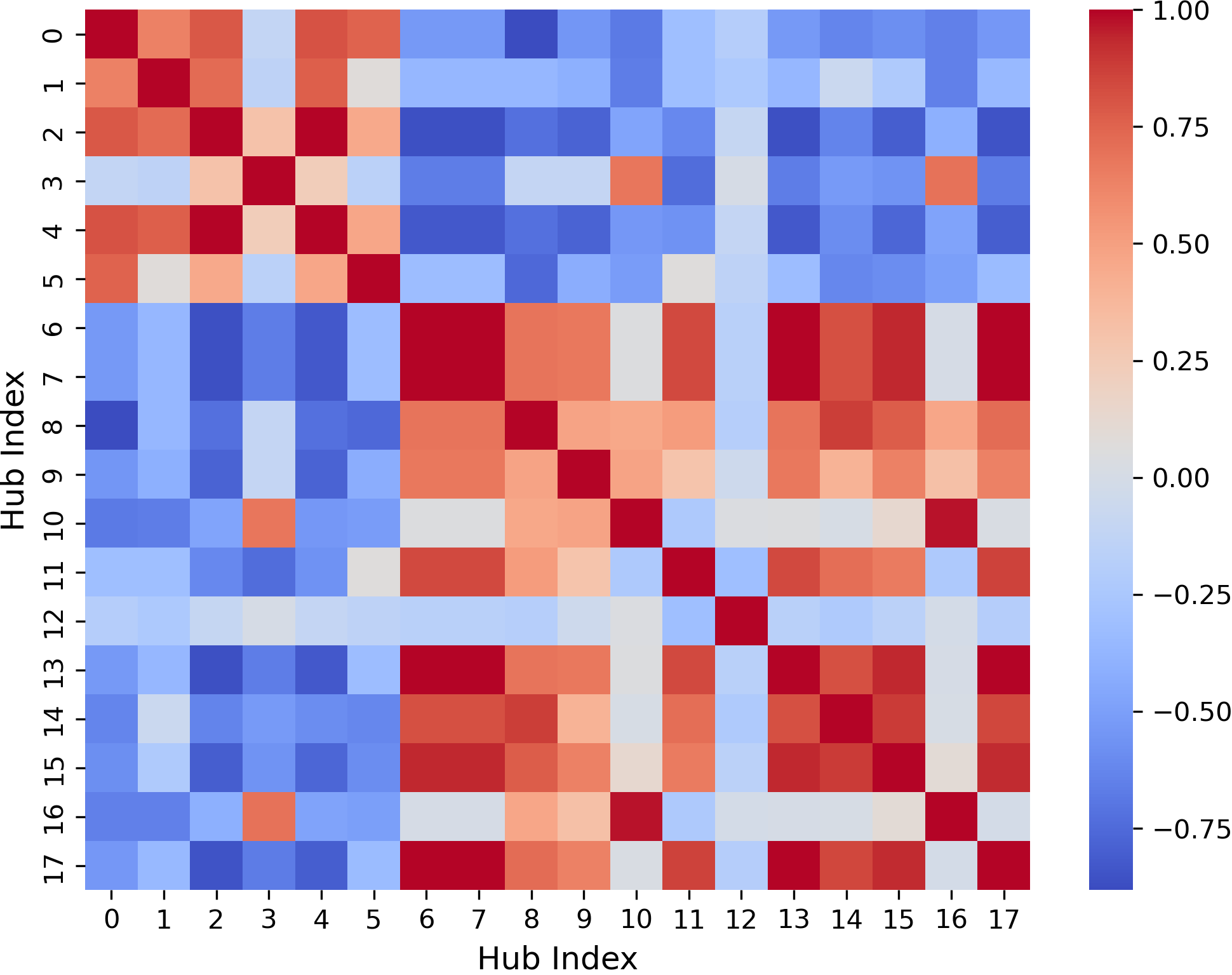}}
    \hfill
    \subfloat[Geographic distribution of hub clusters.]{\includegraphics[height=0.29\textheight,keepaspectratio]{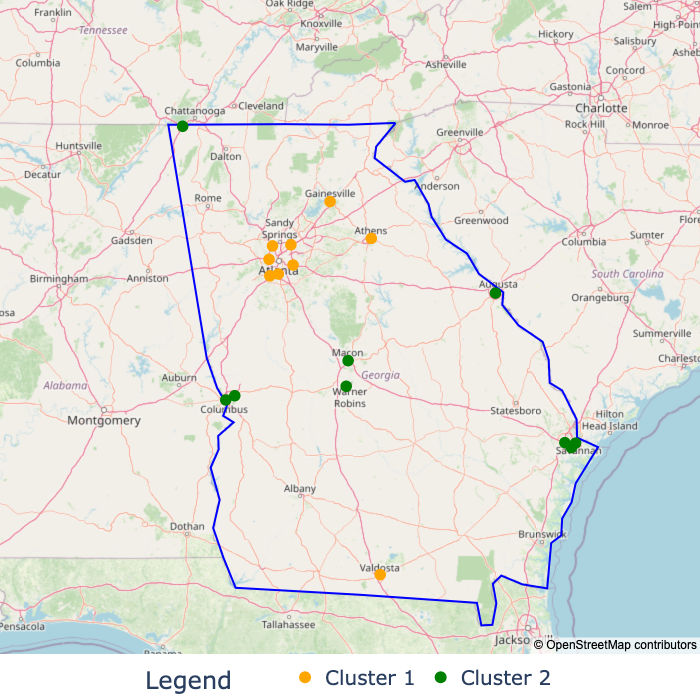}}
    \caption{Risk-based hub similarities and clusters.}
    \label{fig:similarity}
\end{figure}

In the second step, daily risk assessments for each logistic hub in 2024 are generated using the selected tools and top risk types, with the results aggregated into a comprehensive yearly summary. This summary enables an in-depth analysis of how hubs are connected through their risk profiles, uncovering patterns and relationships essential for strategic decision-making. To evaluate the similarity between hubs, a cosine similarity \cite{huang2008similarity} measure is applied to standardized risk profiles, capturing how closely hubs align in terms of exposure to key risks, including Thunderstorm Wind, Tornado, Flash Flood, and Traffic Jam. The resulting similarity matrix is visualized in the heatmap shown in Figure~\ref{fig:similarity}(a), where darker red regions indicate higher similarity between hubs, reflecting shared risk characteristics, while blue regions represent greater disparity in risk exposure.

The clustering analysis, derived from this similarity matrix, further organizes hubs into groups with similar risk profiles. As shown in Figure~\ref{fig:similarity}(b), the geographic distribution reveals two primary clusters. Cluster 1 predominantly consists of hubs in urban areas with high operational risks like traffic jams, suggesting a need for solutions such as advanced traffic monitoring and infrastructure optimization to improve efficiency. Cluster 2, in contrast, is made up of hubs in regions frequently exposed to severe weather events, including tornadoes and flash floods, necessitating robust resilience measures such as real-time weather monitoring, contingency planning, and infrastructure fortification. The alignment between geographic and risk-based clustering has several practical benefits:

\begin{itemize}
    \item \textbf{Regionally Coordinated Risk Management.} Hubs within the same cluster can share resources and collaborate on solutions, fostering more effective regional logistic hub network engineering and management, and enhancing communication between stakeholders. By integrating clustering insights into a resource and risk management map, decision-makers can better visualize vulnerabilities and opportunities, facilitating precise resource allocation and the implementation of targeted interventions.
    
    \item \textbf{Efficient Implementation of Mitigation Measures.} Since hubs within the same cluster are geographically close, mitigation strategies such as real-time weather monitoring and infrastructure upgrades can be deployed more efficiently. This proximity reduces operational costs and prevents unnecessary duplication of efforts.

    \item \textbf{Identification of Unexpected Borders in Risk Exposure.} Risk exposure does not always conform to traditional geographic boundaries. Identifying these unexpected borders helps decision-makers focus on areas that require additional attention or customized mitigation strategies beyond predefined regions.

    \item \textbf{Hub Network Evolution and Reconfiguration.} The clustering results support the continuous evolution of the logistic hub network by enabling dynamic adjustments in response to shifting risks and operational demands. This framework facilitates the strategic reconfiguration of hubs based on evolving risk assessments, ensuring that logistics systems remain efficient, resilient, and adaptable to both current challenges and future uncertainties.

\end{itemize}

This study underscores the value of integrating LLM-driven risk assessment with clustering techniques to enhance logistic hub planning. By systematically identifying risk patterns and grouping hubs with similar profiles, this approach enables more data-driven, adaptive decision-making. The proposed framework provides a foundation for optimizing resource allocation, improving network resilience, and guiding long-term strategic planning. As risks and operational demands evolve, this methodology ensures that logistics systems remain robust, flexible, and capable of addressing future uncertainties in hyperconnected supply chains.

\section{Conclusion}
In this paper, we propose an LLM-driven risk assessment framework for hyperconnected logistic hub network deployment, integrating multiple analytical tools to systematically identify and evaluate risks. By applying similarity analysis and hierarchical clustering, we demonstrate that risk-based hub clusters closely align with geographic clustering, reinforcing the practicality of this approach for strategic decision-making. Our findings reveal that risk exposure does not always conform to traditional geographic boundaries, highlighting the need for adaptive risk management strategies. The integration of real-time data sources and analytical tools enables precise resource allocation, enhances hub resilience, and supports proactive decision-making in complex supply chain environments.

In the future, this work can be extended by incorporating dynamic risk factors, refining clustering methodologies for more precise risk segmentation, and proposing new hub locations based on risk-informed analysis. Additionally, integrating a multi-agent system could enhance scalability and efficiency through parallel analysis and distributed decision-making, further strengthening its applicability in hyperconnected supply chain networks.

\bibliography{references}

\begin{thebibliography}{23}
\providecommand{\natexlab}[1]{#1}
\providecommand{\url}[1]{\texttt{#1}}
\expandafter\ifx\csname urlstyle\endcsname\relax
  \providecommand{\doi}[1]{doi: #1}\else
  \providecommand{\doi}{doi: \begingroup \urlstyle{rm}\Url}\fi

\bibitem[Aljohani(2023)]{aljohani2023predictive}
Abeer Aljohani.
\newblock Predictive analytics and machine learning for real-time supply chain risk mitigation and agility.
\newblock \emph{Sustainability}, 15\penalty0 (20):\penalty0 15088, 2023.

\bibitem[Arowosegbe et~al.(2024)Arowosegbe, Olutimehin, Odunaiya, and Soyombo]{arowosegbe2024sustainability}
Oluwakemi~Betty Arowosegbe, David~Olanrewaju Olutimehin, Olusegun~Gbenga Odunaiya, and Oluwatobi~Timothy Soyombo.
\newblock Sustainability and risk management in shipping and logistics: Balancing environmental concerns with operational resilience.
\newblock \emph{International Journal of Management \& Entrepreneurship Research}, 6\penalty0 (3):\penalty0 923--935, 2024.

\bibitem[Boppiniti(2019)]{boppiniti2019machine}
Sai~Teja Boppiniti.
\newblock Machine learning for predictive analytics: Enhancing data-driven decision-making across industries.
\newblock \emph{International Journal of Sustainable Development in Computing Science}, 1\penalty0 (3), 2019.

\bibitem[Choudhary et~al.(2023)Choudhary, Singh, Schoenherr, and Ramkumar]{choudhary2023risk}
Nishat~Alam Choudhary, Shalabh Singh, Tobias Schoenherr, and M~Ramkumar.
\newblock Risk assessment in supply chains: a state-of-the-art review of methodologies and their applications.
\newblock \emph{Annals of Operations Research}, 322\penalty0 (2):\penalty0 565--607, 2023.

\bibitem[Emrouznejad et~al.(2023)Emrouznejad, Abbasi, and S{\i}caky{\"u}z]{emrouznejad2023supply}
Ali Emrouznejad, Sina Abbasi, and {\c{C}}i{\u{g}}dem S{\i}caky{\"u}z.
\newblock Supply chain risk management: A content analysis-based review of existing and emerging topics.
\newblock \emph{Supply Chain Analytics}, 3:\penalty0 100031, 2023.

\bibitem[Huang et~al.(2008)]{huang2008similarity}
Anna Huang et~al.
\newblock Similarity measures for text document clustering.
\newblock In \emph{Proceedings of the sixth new zealand computer science research student conference (NZCSRSC2008), Christchurch, New Zealand}, volume~4, pages 9--56, 2008.

\bibitem[Jahin et~al.(2023)Jahin, Naife, Saha, and Mridha]{jahin2023ai}
Md~Abrar Jahin, Saleh~Akram Naife, Anik~Kumar Saha, and Muhammad~Firoz Mridha.
\newblock Ai in supply chain risk assessment: A systematic literature review and bibliometric analysis.
\newblock \emph{arXiv preprint arXiv:2401.10895}, 2023.

\bibitem[Kulkarni et~al.(2022)Kulkarni, Dahan, and Montreuil]{kulkarni2022resilient}
Onkar Kulkarni, Mathieu Dahan, and Benoit Montreuil.
\newblock Resilient hyperconnected parcel delivery network design under disruption risks.
\newblock \emph{International Journal of Production Economics}, 251:\penalty0 108499, 2022.

\bibitem[Li et~al.(2024)Li, Liu, Dahan, and Montreuil]{li2024stochastic}
Jingze Li, Xiaoyue Liu, Mathieu Dahan, and Benoit Montreuil.
\newblock Stochastic service network design with different operational patterns for hyperconnected relay transportation.
\newblock \emph{arXiv preprint arXiv:2402.06222}, 2024.

\bibitem[Montreuil(2011)]{montreuil2011toward}
Benoit Montreuil.
\newblock Toward a physical internet: meeting the global logistics sustainability grand challenge.
\newblock \emph{Logistics Research}, 3:\penalty0 71--87, 2011.

\bibitem[Montreuil et~al.(2018)Montreuil, Buckley, Faugere, Khir, and Derhami]{montreuil2018urban}
Benoit Montreuil, Shannon Buckley, Louis Faugere, Reem Khir, and Shahab Derhami.
\newblock Urban parcel logistics hub and network design: The impact of modularity and hyperconnectivity.
\newblock 2018.

\bibitem[Muthukrishnan et~al.()Muthukrishnan, Kulkarni, and Montreuil]{muthukrishnanresilient}
Praveen Muthukrishnan, Onkar~Anand Kulkarni, and Benoit Montreuil.
\newblock Resilient logistics flow routing in hyperconnected networks.

\bibitem[Nezianya et~al.(2024)Nezianya, Adebayo, and Ezeliora]{nezianya2024critical}
Michelle~Chibogu Nezianya, Ahmed~Olanrewaju Adebayo, and Paschal Ezeliora.
\newblock A critical review of machine learning applications in supply chain risk management.
\newblock \emph{World Journal of Advanced Research \& Reviews}, 23\penalty0 (3):\penalty0 1554--1567, 2024.

\bibitem[Pournader et~al.(2020)Pournader, Kach, and Talluri]{pournader2020review}
Mehrdokht Pournader, Andrew Kach, and Srinivas Talluri.
\newblock A review of the existing and emerging topics in the supply chain risk management literature.
\newblock \emph{Decision sciences}, 51\penalty0 (4):\penalty0 867--919, 2020.

\bibitem[Quan and Liu(2024{\natexlab{a}})]{quan2024econlogicqa}
Yinzhu Quan and Zefang Liu.
\newblock Econlogicqa: A question-answering benchmark for evaluating large language models in economic sequential reasoning.
\newblock \emph{arXiv preprint arXiv:2405.07938}, 2024{\natexlab{a}}.

\bibitem[Quan and Liu(2024{\natexlab{b}})]{quan2024invagent}
Yinzhu Quan and Zefang Liu.
\newblock Invagent: A large language model based multi-agent system for inventory management in supply chains.
\newblock \emph{arXiv preprint arXiv:2407.11384}, 2024{\natexlab{b}}.

\bibitem[Sun et~al.(2024)Sun, Wen, Ping, and Zhang]{sun2024application}
Jun Sun, Xin Wen, Gang Ping, and Mingxuan Zhang.
\newblock Application of news analysis based on large language models in supply chain risk prediction.
\newblock \emph{Journal of Computer Technology and Applied Mathematics}, 1\penalty0 (3):\penalty0 55--65, 2024.

\bibitem[Tummala and Schoenherr(2011)]{tummala2011assessing}
Rao Tummala and Tobias Schoenherr.
\newblock Assessing and managing risks using the supply chain risk management process (scrmp).
\newblock \emph{Supply Chain Management: An International Journal}, 16\penalty0 (6):\penalty0 474--483, 2011.

\bibitem[Wang et~al.(2016)Wang, Gunasekaran, Ngai, and Papadopoulos]{wang2016big}
Gang Wang, Angappa Gunasekaran, Eric~WT Ngai, and Thanos Papadopoulos.
\newblock Big data analytics in logistics and supply chain management: Certain investigations for research and applications.
\newblock \emph{International journal of production economics}, 176:\penalty0 98--110, 2016.

\bibitem[Xu et~al.()Xu, Klibi, and Montreuil]{xu5144739dynamic}
Yujia Xu, Walid Klibi, and Benoit Montreuil.
\newblock Dynamic deployment of pooled human-robot resources in urban parcel logistics.
\newblock \emph{Available at SSRN 5144739}.

\bibitem[Yang et~al.(2023)Yang, Lim, Qu, Ni, and Xiao]{yang2023supply}
Mei Yang, Ming~K Lim, Yingchi Qu, Du~Ni, and Zhi Xiao.
\newblock Supply chain risk management with machine learning technology: A literature review and future research directions.
\newblock \emph{Computers \& Industrial Engineering}, 175:\penalty0 108859, 2023.

\bibitem[Yao et~al.(2022)Yao, Zhao, Yu, Du, Shafran, Narasimhan, and Cao]{yao2022react}
Shunyu Yao, Jeffrey Zhao, Dian Yu, Nan Du, Izhak Shafran, Karthik Narasimhan, and Yuan Cao.
\newblock React: Synergizing reasoning and acting in language models.
\newblock \emph{arXiv preprint arXiv:2210.03629}, 2022.

\bibitem[Zhao et~al.(2024)Zhao, Hussain, Zhang, and Saberi]{zhao2024optimizing}
Ming Zhao, Omar Hussain, Yu~Zhang, and Morteza Saberi.
\newblock Optimizing supply chain risk management: An integrated framework leveraging large language models.
\newblock In \emph{2024 IEEE Conference on Artificial Intelligence (CAI)}, pages 1057--1062. IEEE, 2024.

\end{thebibliography}

\end{document}